# Adaptation of pedagogical resources description standard (LOM) with the specificity of Arabic language

Asma Boudhief [#], Mohsen Maraoui [*], Mounir Zrigui [#]

[#] *UTIC laboratory, Higher School of Sciences and Techniques of Tunis, Tunisia*
*UTIC Laboratory, University of Monastir, Tunisia*
[1] asmaboudhief@live.com
[3] maraoui.mohsen@gmail.com

[*] *UTIC Laboratory, University of Monastir, Tunisia*
[2] mounir.zrigui@fsm.rnu.tn

*Abstract*— **In this article we focus firstly on the principle of pedagogical indexing and characteristics of Arabic language and secondly on the possibility of adapting the standard for describing learning resources used (the LOM and its Application Profiles) with learning conditions such as the educational levels of students and their levels of understanding,... the educational context with taking into account the representative elements of text, text length, ... in particular, we put in relief the specificity of the Arabic language which is a complex language, characterized by its flexion, its voyellation and agglutination.**

I. INTRODUCTION

With the advancing technology, the teaching of the language has undergone great changes and teachers use computers to better present their course, while the existing systems for the Arabic language does not meet their needs, there are static systems, characterized by the absence of auto-correction and the absence of changes in exercises in the same unit of learning.

Presenting a course that meets the needs of teachers requires the right choice of text, which seems difficult because of the lack of tools allowing access to the texts according to desired criteria.

"*Although the text search seems to be a recurring tasks in language teaching, it seems that few tools have been designed to enable teachers to access texts based on criteria related to their problems*" [1].

Exists Mechanisms of search are based on a traditional search by keywords and this mechanism seems inefficient, it requires a pedagogic text's indexing to facilitate his search.

II. ETAT DE L'ART

A. *Model of information system*

"*An IRS (Information Retrieval System) is a computerized system that facilitates access to a set of documents (corpus), to help find those whose content best fits for need of information of a user.*" [2].

In an IRS (see Fig. 1) we perform in the first hand, an indexing of existing documents in the database to obtain a model of documents. On the other hand if a user sends a request, it will be interpreted and the system creates facets representing this query. Then it performs a match with the model of documents to extract the most relevant documents at the request of the user.

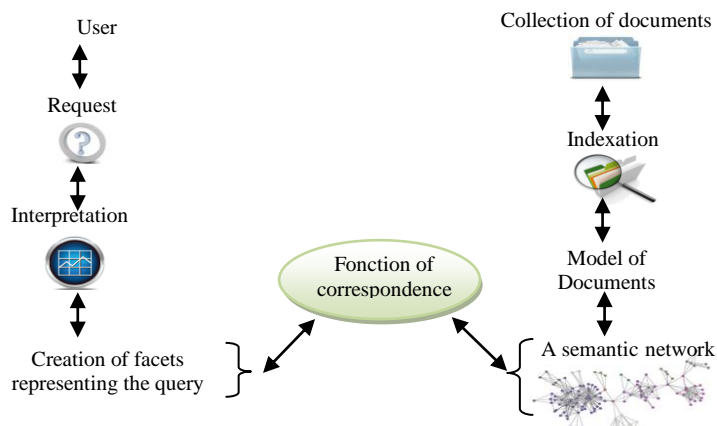

Fig. 1: scheme of a model of information retrieval

For each system, you must:
- Define what type of documents will be processed. i.e: paragraph, text page or multiple pages.
- Establish an indexing by extracting words deemed relevant in defining the document in order to obtain templates.
- Creation of semantic networks to present the document templates

On the other hand, a request coming from the user should be construed to create representative facets, and then the system establishes the correspondence between the created facets and models of the semantic network by calculating the degree of similarity [3] and releasing the document nearest to the user's request.

## B. Pedagogical indexing

According to Loiseau, pedagogical indexing [4] is an "indexing performed according to a documentary language that allows the user to search for objects to use in education"[1].
Our thesis aims to propose a model for what we called the indexation of pedagogical texts for teaching the Arabic language and demonstrate its feasibility by implementing a prototype.

This leads us to insist that such a database must allow the following use cases:
- ➢ adding text to the base
- ➢ Text searching based on the problematic issues and specificity of the Arabic language.
- ➢ Aid for selecting text

The indexing operation consists of analyzing the object to be indexed by "extracting concepts", and finally express it in a documentary language.
The agent of these operations is not specified. So we can imagine several configurations: the analysis can be performed by a human operator or a machine, and the expression of concepts extracted in documentary language can be performed either by a human or a machine [1].

In our case, the indexing operation will involve the user and the system. The user is not a documentalist and as a teacher, it is primarily the use case "text's search" which interests him.
Both sides of the indexing process will therefore be as simple and not boring as possible and to do that, they must be automated as possible.

The analysis of some concepts of the document cannot be automated, such as the author or title (if these criteria are relevant to their operation in language teaching). But any automated analysis must be supported by the system. The analysis part of the indexing will be hybrid in the sense that some concepts cannot be managed by the system, but the most fastidious will be automated where possible.

In what follows, we will explain the influence of pedagogical context on the choice of the text, before exploring the existing standard description of educational resources so that we may adapted to the specificity of the Arabic language and needs users, and then introduce the notion of facet of a text and present a model in which it occurs.

## C. Influence of pedagogic context in the choice of the text

After the formulation of the problem (setting up an activity), a text was assigned successively projected properties. We call these properties specified at progressively steps, the learning environment.
The pedagogical context is defined as "the set of features describing the teaching situation"[5].

As part of the educational indexing, it is not appropriate to set all properties that may intervene in the educational context (CP), but rather to try to identify the relevant components of the text search as shown in Fig. 2.

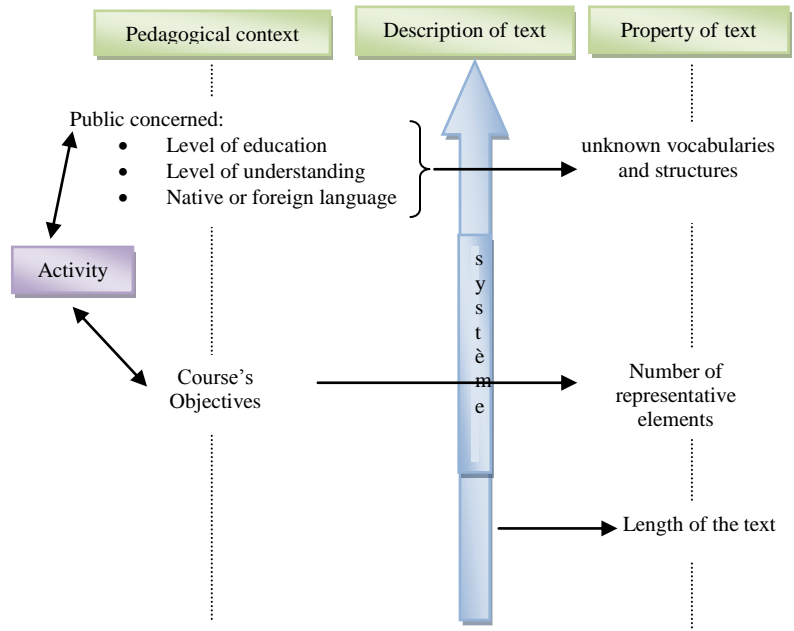

Fig. 2 Effect of pedagogical context on assigning properties to the text

According to the diagram, we see that the assignment of properties to the text depends to:
- The educational context: he relates, firstly the main objectives of the course (during conjugation of verbs, grammar courses ...) and secondly the targets publics as the level of students, their understanding and the studied language (native or foreign).
- The description of the text: which we can determine the length of the text which is an important criterion of language education, and its title.

From these two criteria, we can determine some properties of text such as:
- ❖ Its length: the majority of language teachers want to have a significant text and minimum length to not annoy the learner.
- ❖ The representative elements: for example if the student requests an exercise for the conjugation of an Arabic verb composed of three letters in the past, our representative element is the verb.

❖ The unknown vocabularies for students: for example if the student knows the verbs composed of three letters in Arabic but he does not know the conjugation of its verbs "mootalla" they contain the letter "alif" (أ), so we must consider this case.

### III. USER'S NEEDS

In order to define appropriate criteria for indexing pedagogical text for teaching the Arabic language, we carried out a brief questionnaire to understand how teachers classify their texts and to locate the search criteria

*A. The questionnaire results*

In our questionnaire we asked a group of teachers (60 teachers, fewer but just to know the opinion of a community of teachers) whose teach students ranging in age from 7 years and 18 years, that is to say, from primary to secondary level.
According to the questionnaire teachers of Arabic say "be able to use the same text in several different contexts."
We noticed that they have recourses specifically to build their own texts when they want to control their content language (grammatical structures, vocabulary), so there is no teacher that automatically searches for a text, but rather depends on teaching conditions (the objective of the course, student level ...).
With the description of the teacher concerning the classification of its own collection of text, we can isolate certain search criteria, the most widely used were: the purpose, content and level.
The analysis results also allowed us to deduce that the majority of teachers have obtained texts when searching for a specific activity, and then come the predefined text in the program.
Another, teachers focus mainly on certain statements in the context of the lesson, then on the subject and the length of the text and especially all the teachers are trying to choose a text of length as a minimum.

### IV. GRAMMAR, COMPOSITION AND CHARACTERISTICS OF ARABIC LANGUAGE

Arabic is a language that:
  ➢ The words are written in horizontal lines from right to left.
  ➢ Most letters change shape depending on whether they appear at the beginning, middle or end of a word.
  ➢ The letters can be joined are always united both handwritten and printed.
  ➢ The vowels are added below and above the letters.

*A. Composition of arabic language*

In Arabic, a sentence can be either nominal "إسمية" or verbal "فعلية".
A nominal phrases is composed of two parts:
  ➢ « al mobtada » : In its simplest form is a unique and determined name. Example: المطرُ غزيرٌ (**the rain** is pouring).
    But it may take other complex shapes as: « mourakab naati » : الْأَشْجَارُ الْخَضْرَاءُ نَافِعَةٌ (**green trees** are useful)
  ➢ « al khabar » : In its simplest form is a single name. Example : المطرُ غزيرٌ (the rain **is pouring**)

A verbal clause constitutes either of a verb + subject or a verb + subject + complement (object, time or place):

  ➢ Verb: In viewpoint tenses of conjugation, the verb in Arabic is composed of three types: past tense, future tense or order.
    On the conjugation of verbs you can find a common structure between the verbs in the same category.
    Example: Verbs "thoulethi moujarrad" in Past and future

    فَعَلَ- يَفْعُلُ : نَصَرَ - يَنْصُرُ
    فَعَلَ- يَفْعِلُ : جَلَسَ - يَجْلِسُ
    فَعَلَ- يَفْعَلُ: مَنَعَ - يَمْنَعُ
    فَعَلَ- يَفْعَلُ: عَلِمَ - يَعْلَمُ
    فَعَلَ- يَفْعِلُ: حَسِبَ - يَحْسَبُ
    فَعَلَ- يَفْعُلُ: كَرُمَ – يَكْرُمُ

  ➢ Subject: It is a name "marfoua" that precedes the verb which means who release the action.
  ➢ Object complement: It's over which it exercises the action, it is always "Mansoub" and can be explicit or implicit.

*B. Property of the arabic language*

The Arabic language has the following properties:

➢ Inflected language :
  It is a language in which lexical units vary in number and in bending (the number of names, or verb tense) according to the grammatical relationships they have with other lexical units.

kataba AlAwlAd+u        كتب الأولادُ
(V)PAST (N)+NOM
 Wrote  kids
"The kids wrote"
qAbAlA samir AlAwlAd+a   قابل سمير الأولادَ
(V)PAST (N) (N)+ACC
met samir childrens
"samir met childrens"

salama samir 3ly AlAwlAd+i   سلّم سمير على الأولادِ
(V)PAST (N) (PREP) (N)+GEN
welcomed samir children
"samir welcomed children"

- The voyellation :
  An Arabic lexical unit is written with consonants and vowels. Vowels are added above or below letters. They are required for reading and for a correct understanding of a text and they can differentiate lexical units having the same representation.

- The agglutination :
  The Arabic language is agglutinative that clitics stick to nouns, verbs, adjectives which they relate. These phenomena pose formidable problems for the automatic analysis of Arabic, as in so far as that they greatly increase the rate of ambiguity by introducing additional ambiguities in the segmentation of words. Indeed, an Arabic word may have several possible divisions: proclitic, flexive form and enclitic.

- Pro-drop (=to an empty pronominal subject) :
  The ASM neglects systematically the morphological realization of subject pronoun. However, the verb agrees in person, number and gender with the pronoun omitted, as the following example shows: (The pronoun call is placed between brackets).

*akaluu {humu}*          vs    *akalnna {hunna}*
(V)PAST.3.MASC.PL              (V)PAST.3.FEM.PL
have eaten {they}              have eaten {they}
'they have eaten.' (أكلوا)     'they have eaten.' (أكلن)[6]

C. *Recent classification of arabic units*

A classification is fairly recent one made by Khoja[7] in the development of a morphosyntactic tagger.
Khoja presents a label based on the traditional classification and refined by the subdivisions proposed by Haywood [8].
Under this classification, the lexical units are divided into five classes: noun, verb, particle, residual and punctuation. Some are refined into sub classes illustrated in Fig.3.

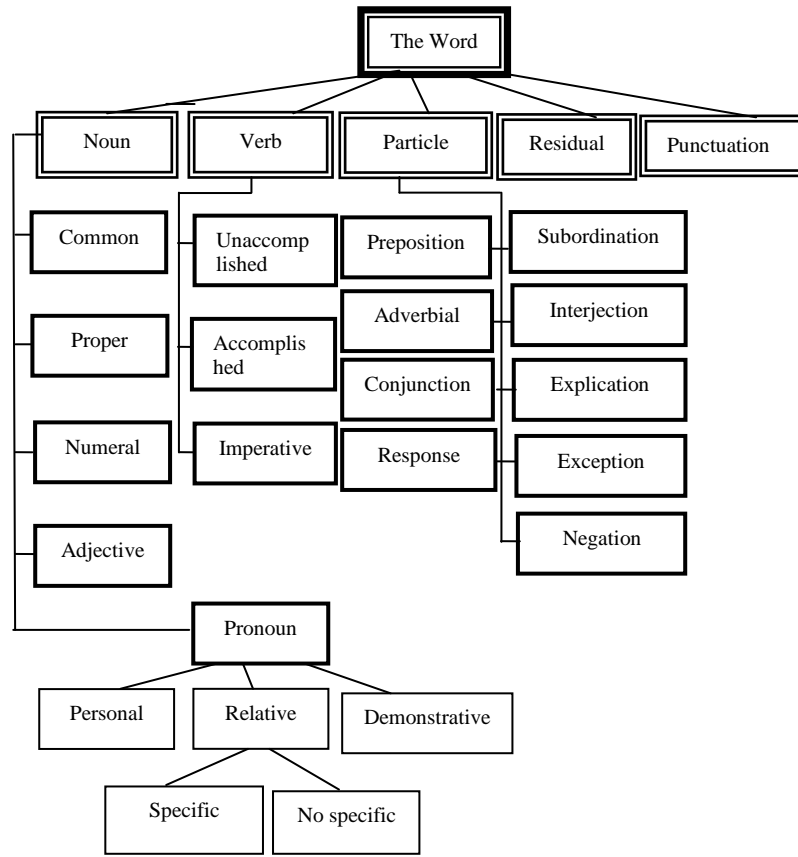

Fig. 3 Classification of Arabic lexical units

V. DESCRIPTION OF PEDAGOGICAL RESOURCES

The indexing of learning objects is an absolute necessity if we want to find them. For that, you have to add semantic information. This information is metadata: data describing data [9].

For that metadata fulfill their role and facilitate access to online resources, it is imperative that a stable standard that exists to providers of resource and users can use the same repository. This standard must also support developments and extensions to accommodate new needs.

A. *The description's standard of pedagogical resources (LOM)*

*The Learning Object Metadata* (LOM) is a standard published in 2002 by *the Learning Technology Standards Commitee* (LTSC) de l'IEEE (*Institute of Electrical and Electronics Engineers*). The standard consists of four parts:
• IEEE 1484.12.1 - Conceptual model of metadata;
• IEEE 1484.12.2 - Implementation of ISO / IEC 11404 in the LOM metadata model;
• IEEE 1484.12.3 - Development and implementation of the XML Schema for LOM;

• IEEE 1484.12.4 - Definition of application framework RDF (Resource Description Framework) for LOM [10].

All LOM elements are optional, that is to say that the model can work without all the fields are filled. Nevertheless, it is desirable to provide the most information so that resources can be maximum exploited.

The LOM is organized into nine categories performing different functions. The elements contained in each category can be seen in Fig. 4

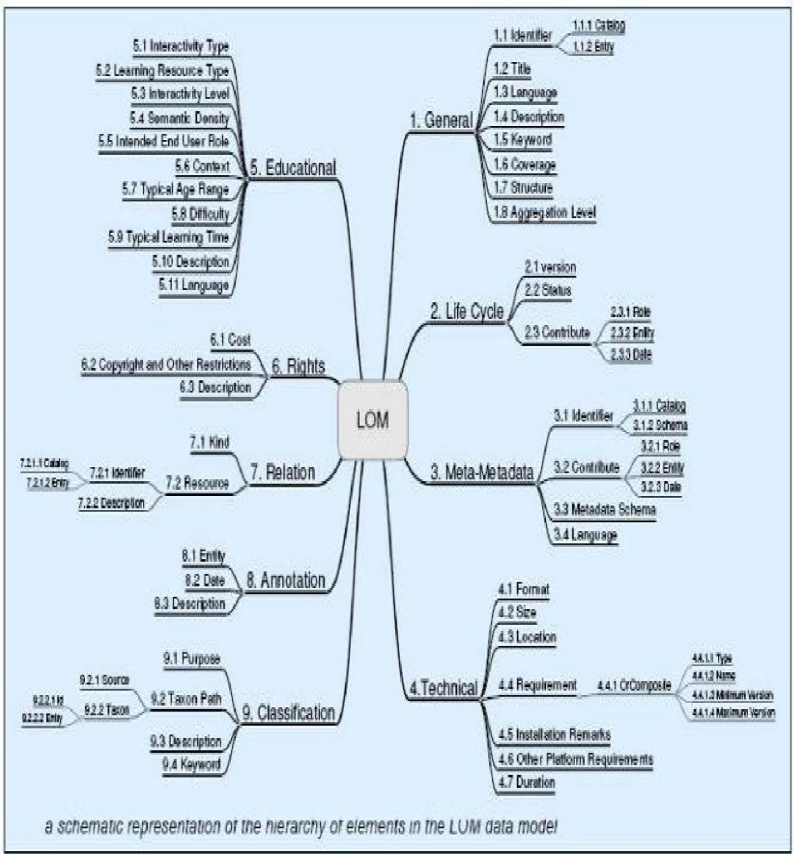

Fig. 4: Organization of LOM metadata scheme [11]

The LOM standard is:
- Fairly complex because it includes 78 elements forming a tree on three floors and offers 59 fields of information independent.
- An Abstract model which must be instantiated in a particular context. It is therefore to adapt this standard to respond to specific and real needs of users.

B. *Conformity between the structure adapted to the Arabic language and the standard LOM*

In our system we treated three levels depending on the degree of complexity of the components of the sentence:

❖ Level 1: treats the notion of verb, noun and particle:
  ✓ The verb: We treat the notion of the verb without indicate his form, type or time in which is introduced
  ✓ Name: This is the notion of a name devoid of its functionality in the sentence
  ✓ The particle: This is the notion of a particle in general, without detailed its category (preposition, coordination, demonstrative, relative pronoun, ...)

❖ Level 2: In which we treat the components of a verbal phrase (verb, subject, object) and the components of a nominal phrase ("mobtada" and "khabar") and other concepts such as the complement of place, the complement of time, ...

❖ Level 3: At this level we treat the concept of composite words ("mourakeb jar", "morakeb atfi", "mourakeb idhafi", ...)

In order to the system to meet the needs of the user (teacher or learner) and from the chosen level, we must store information about the texts used for teaching the Arabic language in an XML file (Extensible Markup Language). This file should follow the elements of the LOM.

Note that for compliance with the LOM, it is not mandatory to complete all elements LOM, empty LOM instance is conforming.

So we could reuse the LOM although some elements (Fig. 4) are not relevant in the context of our pedagogical work of indexing.

In this context, we focus on the category 5 "educational" that interests us for our indexing problem.

This category describes the key educational or educational characteristics of the learning object and it contains 11 elements located below.

❖ **Interactivity Type:** It's the type of interaction between the resource and the user (Active, Expositive, Mixed)
❖ **Learning Resource Type:** It's the type teaching (exercise, simulation, questionnaire, diagram, figure, graph, index, slide, table, narrative text, exam, experiment, problem statement, self assessment, presentation)
❖ **InteractivityLevel:** degree of interactivity which can be set very low, low, medium, high, very high.
❖ **SemanticDensity:** which also takes values very low, low, medium, high, and very high.
❖ **Intended end user role:** a resource user is in our case a teacher or learner.
❖ **Context:** the environment of resource's use (school, higher education, training, other)

- **Typical Age Range:** age of the user
- **Difficulty:** difficulty of the resource (very easy, easy, medium, difficult, very difficult)
- **Typical Learning Time:** Approximate or typical time to work with the resource
- **Description:** Comments on the use of the resource, in which we will detail the composition of sentences Arabic texts used
- **Language:** the language of the user who is in our case the Arabic language

**Description** is the element through which we can indicate that a text can be used in different contexts. Each context will match a record with the values of other fields. So in this element we provide a standardized schema to provide information about the grammatical functionalities of the text which can be retrieved by the system in response to a specific need of the user.

And since the LOM standard does not require the integration of fields describing the grammatical functionalities of the European language, so we can describe fields similar to the European language such as verbs, nouns, and other fields ... non-existent in that language as the words composed ("mourakeb jar, mourakeb idhafi")

## VI. LOOKING FOR A TEXT IN AN INDEXED TEXT BASE

The primary function of a text database indexed for language teaching is to allow the search of text.
In our system, we defined the modalities of interaction between such a system and a user-student, or such a system and a user-teacher looking for a text that meets the needs of each with using the concept of facet-prism.
The texts used are encoded using UTF-8. The choice of this coding was justified by the fact that, firstly, most Arab digital textual resources are encoded using this standard, and secondly because the standard UTF-8 is supported by most popular browsers.

### A. The research of activity for a student

After his authentication, the student is faced with an interface representing the types of exercises available on:
- Morphology (الصرف): conjugation with past, future... with verbs of different classes
- The sentence composition and role of their components

The exercises are characterized by levels of difficulty depending on the complexity of the composition of the sentence and the complexity of verb conjugation. (Fig.5)

Fig.5: interface representing the menu displayed for a student

Once a student specifies the pedagogical context (CP), the system performs its processing:
- It calculates the necessary facets
- It extracts using facets computed before, the collection C corresponding to CP

In the following step:
- The system offers the student an exercise E1 of the collection C
- The student responds to exercise
- The system performs the comparison process between the student's response and the response existing in the database and then displays the correction, by characterizing the wrong answers with red color and the correct answers by the color green.

If the student wants to try another exercise in the same class, he asks the system, the latter displays a new exercise E2∈C\{E1}, and so on until the student's request for stop the process. Our system contains a variety of types of exercises:
- Text hole: in this type of exercise, the system gives either the set of words to fill (fig.6), either in each empty zone, the system provided a set of proposal as a "text select" (fig.7), these proposals are selected depending on the type of the correct answer, i.e. if we have a conjunctive preposition type response, the system gives the learner a list of prepositions of this type, which makes our system more dynamic.
- Multiple choices (fig.8)
- Question / Answer (fig.9)
- ... Etc.

Fig.6: Interface representing an exercise with type text blanks

Fig.7: Interface representing another exercise with type text blanks

Fig.8: Interface representing an exercise with type multiple choice question

Fig.9: Interface representing an exercise with type question / response

B. THE RESEARCH OF TEXT FOR A TEACHER

In the case where a teacher wants to search a text for a specific activity, it must specify the intended pedagogical context (CP), ie category and level of complexity of exercises. Then the system:

- calculates the values of facets related to CP
- Seeking the collection C in the base of text
- Presents the collection C to the teacher
- As soon as the collection C is ready a new interface is displayed containing a list of texts that meet the requirements of the teacher with a drop-down list containing the script that a teacher can perform on selected text.

Right now, teachers are asked to determine the number of the selected text $t_i$ with the script he wants to apply (Fig.10).

A new interface will appear containing the exercise resulting from the combination of the script with the selected text and so on until the request of stopping the process.

Fig.10: interface representing the choice of a text and the type of exercise by a teacher

## VII. CONCLUSION

Based on the analysis we made, we find that the Arabic language is a complex language characterized by its richness, flexion, voyellation and its agglutination, which requires a good knowledge of the language so that we can realize a dynamic system that meet the needs of users (teacher or learner).

This implementation is based on the idea of pedagogical indexing and finding data structure texts according to the standard description of educational resources (LOM). But according to our studies on this standard we found that on the one hand all the elements of this standard are optional and on the other hand, the class "educational" (the category that interests us in the indexing problem) has an element "description" in which we can integrate the fields describing a text without to interfere with the constraints of the Arabic language.

REFERENCES

[1] M. Loiseau, G. Andoniadis, , and C. Ponton, "Pratiques enseignantes et « contexte pédagogique » dans le cadre de l'indexation pédagogique de textes," *Congrès Mondial de Linguistique Française - CMLF 2010*, Juil. 2010.
[2] M. Géry, "Indexation et interrogation de chemins de lecture en contexte pour la recherche d'information structurée sur le web," Ph.D. thesis, University Joseph Fourier - Grenoble I, Oct. 2002.
[3] R. Ayadi, M. Maraoui, and M. Zrigui, "Intertextual distance for Arabic texts classification," *The 4th International Conference for Internet Technology and Secured Transactions (ICITST-2009)*, Nov. 2009.
[4] M.A. Ben Mohamed, D. El Ghoul, M.A. Nahdi, M. Mars and M. Zrigui, "Arabic Call system based on pedagogically indexed text," *The 2011 International Conference on Artificial Intelligence (ICAI'11), 2011*, WORLDCOMP'11, 2011.
[5] M. Maraoui, G. Antoniadis, and M. Zrigui, "CALL System for Arabic Based on Natural Language Processing Tools," *The 4th Indian International Conference on Artificial Intelligence, IICAI 2009*, Dec, 2009.
[6] S. Boulaknadel, "Apport des connaissances morphologiques et syntaxiques pour l'indexation," Ph.D. thesis, University of Nantes, Oct. 2008.
[7] S. Khoja, " Thematic Indexing in Video Databases," Ph.D. thesis University of Southampton, United Kingdom, Jan. 2001.
[8] J. A. Haywood and H. M. Nahmad*, A new Arabic grammar*, Lund Humphries Publishers Ltd. 2nd ed. London : Percy Lund Humphries Publishers Ltd, 1965.
[9] O. Lassila, "Web Metadata: A Matter of Semantics," *IEEE Internet Computing*, pp. 30–37, Juil-Aout. 1998.
[10] R.M Gomez de regil, "Présentation des standards : (LOM) – Learning Object Metadata," *enssib, Villeurbanne,* 2004.
[11] (2012)The sticef website. [Online]. Available: http://sticef.univ-lemans.fr/num/vol2004/passardiere 11/sticef_2004_passardiere_11.htm